\documentclass{article} 
\usepackage{fullpage,times}

\usepackage{amsmath,amsthm,amssymb,graphicx,wrapfig}

\usepackage{tikz}
\usetikzlibrary{decorations.pathmorphing,backgrounds,fit}

\newcommand{\eqdef}{\stackrel{\mathrm{def}}{=}}
\newcommand{\err}{\mathrm{err}}

\newcommand{\reals}{\mathbb{R}}
\newcommand{\Y}{\mathcal{Y}}
\newcommand{\X}{\mathcal{X}}
\newcommand{\Z}{\mathcal{Z}}
\renewcommand{\H}{\mathcal{H}}
\newcommand{\E}{\mathbb{E}}
\newcommand{\D}{\mathcal{D}}
\newcommand{\U}{\mathcal{U}}
\newcommand{\inner}[1]{\langle #1 \rangle}

\newcommand{\e}{{\mathbf e}}
\renewcommand{\r}{{\mathbf r}}
\newcommand{\s}{{\mathbf s}}
\newcommand{\x}{{\mathbf x}}

\newcommand{\z}{{\mathbf z}}
\newcommand{\w}{{\mathbf w}}
\newcommand{\T}{\mathrm{time}}

\newcommand{\sig}{\mathrm{sig}}
\newcommand{\poly}{\mathrm{poly}}

\newtheorem{definition}{Definition}
\newtheorem{lemma}{Lemma}

\newtheorem{theorem}{Theorem}

\renewcommand{\eqref}[1]{Eq.~(\ref{#1})}
\newcommand{\lemref}[1]{Lemma~\ref{#1}}

\newcommand{\indct}[1]{\boldsymbol{1}[{#1}]}

\title{Using More Data to Speed-up Training Time}
\date{}

\author{
Shai Shalev-Shwartz \\
The Hebrew University\\
\texttt{shais@cs.huji.ac.il} \\
\and
Ohad Shamir \\
Microsoft Research New-England \\
\texttt{ohad@microsoft.com} \\
\and
Eran Tromer \\
Tel-Aviv University\\
\texttt{tromer@cs.tau.ac.il} \\
}

\begin{document}

\maketitle
\vspace{-0.8cm}
\begin{abstract}
In many recent applications, data is plentiful. By now, we have a
rather clear understanding of how more data can be used to improve
the accuracy of learning algorithms. Recently, there has been a
growing interest in understanding how more data can be leveraged to
reduce the required training runtime. In this paper, we
study the runtime of learning as a function of the
number of available training examples, and underscore the main
high-level techniques. We provide some initial positive results
showing that the runtime can decrease exponentially while only
requiring a polynomial growth of the number of examples, and
spell-out several interesting open problems.
\end{abstract}

\section{Introduction}

Machine learning are now prevalent in a large range of
scientific, engineering and every-day tasks, ranging from analysis of
genomic data, through vehicle and aircraft control to locating
information on the web and providing users with personalized
recommendations.  Meanwhile, our world has
become increasingly ``digitized'' and the amount of data available for
training is dramatically increasing. By now, we have a rather clear
understanding of how more data can be used to improve the
\emph{accuracy} of learning algorithms. In this paper we study how
more data can be beneficiary for constructing more \emph{efficient}
learning algorithms.

Roughly speaking, one way to show how more data can reduce
the training runtime is as follows. Consider learning by finding a
hypothesis in the hypothesis class that minimizes the training
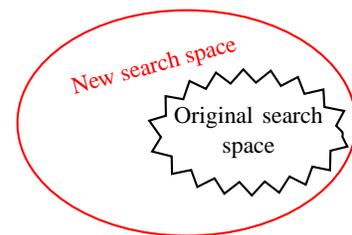
\begin{wrapfigure}{r}{0.3\textwidth}
\begin{center}
\begin{tikzpicture}[scale=0.25]

\draw[thick,black] decorate[decoration=zigzag] {(0,-0.5) ellipse (5 and 3)
};
\node[thick,text width=2cm] at (0,-0.5) {\begin{center} \small Original
    search space
  \end{center}
};

\draw[thick,red] (-3.3,0) ellipse (9 and 6) ;
\node[thick,red,rotate=15] at (-5,3)  {\small New search space};

\end{tikzpicture}
\end{center}
\caption{\small The Basic Approach} \label{fig:illustrate}
\end{wrapfigure}
error. In many situations, this search problem is computationally
hard. One can circumvent the hardness by replacing the original
hypothesis class with a different (larger) hypothesis class, such that
the search problem in the larger class is computationally easier
(e.g., the search problem in the new hypothesis class reduces to a
convex optimization problem). On the flip side, from the statistical
point of view, the estimation error in the new hypothesis class might
be larger than the estimation error in the original class, and thus,
with a small number of examples, learning the larger class might lead
to overfitting even though the same amount of examples suffices for
the original hypothesis class. However, having more training
examples keeps the overfitting in check. In particular, if the number
of extra examples we need for learning the new class is only
polynomially larger than the original number of examples, we end up
with an efficient algorithm for the original problem. If, however, we
don't have those extra examples, our only option is to learn the
original hypothesis class, which may be computationally harder.

The goal of this paper is to present a formal model for studying the
runtime of learning algorithms as a function of the available number
of examples. After defining the formal model, we present a binary classification learning problem for which we can provably (based on standard cryptographic assumption) demonstrate an inverse dependence of the runtime on the number of examples. While there have been previous constructions which demonstrated a similar phenomenon, assuming the existence of a ``perfect'' hypothesis, we show this in the much more natural agnostic model of learning. A possible criticism is that our learning problem is still rather synthetic. We continue with presenting several learning problems, which arise in natural settings, that have more efficient algorithms by relying on the availability of more training data. Some of these examples are based on the intuition of Figure \ref{fig:illustrate}, but some are also based on other ideas and techniques. However, for all these problems, the analysis is based on upper bounds without having matching lower bounds. This raises several interesting open problems.

\subsection{Related Work}\label{subsec:related}

\cite{DecaturGoRo98} were the first to
jointly study the computational and sample complexity, and to show
that a tradeoff between runtime and sample size exists. In particular,
they distinguish between the information theoretic sample complexity
of a class and its computational sample complexity, the latter being
the number of examples needed for learning the class in polynomial
time. They presented a learning problem which is not efficiently learnable from a small training set, and is efficient learnable from a polynomially larger
training set. \cite{Servedio00} showed that for a concept class composed of $1$-decision-lists over $\{0,1\}^n$, which can be learned
inefficiently using $O(1)$ examples, no algorithm can learn it
efficiently using $o(n)$ examples, and there is an efficient algorithm using $\Omega(n)$ examples. The construction was also extended to $k$ decision-lists, $k\geq 1$. with larger gaps.

In contrast to \cite{DecaturGoRo98,Servedio00}, which focused on learning under the realizable case (namely, that the labels are generated by some hypothesis in the class), we mostly focus on the more natural \emph{agnostic} setting, where any distribution over the example domain is possible, and there may be
no hypothesis $h$ in our class that never errs. This is not just a formality
- in both \cite{DecaturGoRo98,Servedio00}, the construction crucially
relies on the fact that the labels are provided by some hypothesis in
the class. In terms of techniques, we rely on the cryptographic
assumption that one-way permutations exist, which is the same
assumption as in \cite{Servedio00} and similar to the assumption in
\cite{DecaturGoRo98}. We note that cryptographic assumptions are common in
proving lower bounds for efficient learnability, and in some sense they
are even necessary \cite{ApplebaumBaXi08}. However, our
construction is very different. For example, in both
\cite{DecaturGoRo98,Servedio00}, revealing information on the identity of the
``correct'' hypothesis is split among many different
examples. Therefore, efficient learning is possible after sufficiently
many examples are collected, which then allows us to return the
``correct'' hypothesis. In our agnostic setting, there is no ``correct''
hypothesis, so this kind of approach cannot work. Instead, our
efficient learning procedure computes and returns an improper
predictor, which is not in the hypothesis class at all.

A potential weaknesses of our example, as well as the example given in
\cite{DecaturGoRo98}, is that our hypothesis class does not consist of ``natural'' hypotheses. The class employed in \cite{Servedio00} is more natural, but it is also a very carefully constructed subset of
decision lists. The goal of the second part of the paper is to
demonstrate gaps (though based on upper bounds) for natural learning
problems.

Another contribution of our model is that it captures the exact
tradeoff between sample and computational complexity rather then only
distinguishing between polynomial and non-polynomial time, which may
not be refined enough. Bottou and Bousquet~\cite{BottouBo08}
initiated a study on learning in the \emph{data laden domain} -- a
scenario in which data is plentiful and computation time is the main
bottleneck. This is the case in many real life applications
nowadays. Shalev-Shwartz and Srebro \cite{ShalevSr08} continued this
line of research and showed how for the problem of training Support
Vector Machines, a joint statistical-computational analysis reveals
how the runtime of stochastic-gradient-descent can potentially
\emph{decrease} with the number of training examples. However, this is only demonstrated via upper bounds. More importantly, the advantage of having more examples only improves running time by constant factors. In this paper, we will be interested in larger factors of improvement, which scale with the problem size.

\section{Formal Model Description}

We consider the standard model of supervised statistical learning,
in which each training example is an instance-target pair and the goal of
the learner is to use past examples in
order to predict the targets associated with future instances. For
example, in spam classification problems, an instance is an email
message and the target is either $+1$ ('spam') or $-1$ ('benign'). We
denote the instance domain by $\X$ and the target domain by $\Y$. A
prediction rule is a mapping $h : \X \to \Y$.  The performance of a
predictor $h$ on an instance-target pair, $(\x,y) \in \X \times \Y$,
is measured by a loss function $\ell(h(\x),y)$. For example, a
natural loss function for classification problems is the 0-1 loss,
$\ell(h(\x),y) = 1$ if $y \neq h(\x)$ and $0$ otherwise.

A learning algorithm, $A$, receives a training set of $m$ examples, $S_m
= ((\x_1,y_1),\ldots,(\x_m,y_m))$, which are assumed to be sampled
i.i.d. from an unknown distribution $\D$ over the problem domain $\Z \subseteq \X \times \Y$.
Using the training data, together with any prior knowledge or
assumptions about the distribution $\D$, the learner forms a
prediction rule. The predictor is a random variable and we
denote it by $A(S_m)$. The goal of the learner is to find a prediction rule
with low generalization error (a.k.a. risk), defined as the expected
loss:
\[ \err(h) ~\eqdef~ \E_{(\x,y) \sim \D}[\ell(h(\x),y)] ~. \]
The well
known no-free-lunch theorem tells us that no algorithm can minimize
the risk without making some prior assumptions on $\D$. Following
the agnostic PAC framework, we require that the learner will find a
predictor whose risk will be close to $\inf_{h \in\H} \err(h)$, where
$\H$ is called a hypothesis class (which is known to the learner).

We use $\err(A(S_m))$ to denote the expected risk
of the predictor returned by $A$, where expectation is with respect to
the random choice of the training set. We denote by $\T(A,m)$ the upper bound on the expected runtime\footnote{To prevent trivialities, we also require
  that the runtime of applying $A(S_m)$ on any instance is at most $\T(A,m)$.} of the algorithm $A$ when running on any training set
of $m$ examples. The main mathematical object that we propose to study
is the following:
\begin{equation}
T_{\H,\epsilon} (m) = \min \{t : \exists~A ~\text{s.t.}~ \forall~\D,~ \T(A,m) \le t \land \err(A(m)) \le \inf_{h \in \H} \err(h) +
\epsilon\} ~,
\end{equation}
where when no $t$ satisfies the above constraint we set
$T_{\H,\epsilon}(m) = \infty$.  Thus, $T_{\H,\epsilon}(m)$ measures
the required runtime to learn the class $\H$ with an excess error of
$\epsilon$ given a budget of $m$ training examples.  Studying this
function can show us how more data can be used to decrease the
required runtime of the learning algorithm. The minimum value of $m$
for which $T_{\H,\epsilon}(m) < \infty$ is the information-theoretic
sample complexity. This corresponds to the case in which we ignore
computation time.  The other extreme case is the value of
$T_{\H,\epsilon}(\infty)$. This corresponds to the \emph{data laden
  domain}, namely data is plentiful and computation time is the only
bottleneck.

We continue with few additional definitions. In general, we make no
assumptions on the distribution $\D$. However, we sometime refer to
the realizable case, in which we assume that the distribution $\D$
satisfies $\min_{h \in \H} \err(h) = 0$. The empirical error on the
training examples, called the training error, is denoted by $\err_S(h)
~\eqdef~ \frac{1}{m} \sum_{i=1}^m \ell(h(\x_i),y_i) $.  A common
learning paradigm is Empirical Risk Minimization, denoted ERM$_\H$, in
which the learner can output any predictor in $\H$ that minimizes
$\err_S(h)$. A learning algorithm is called \emph{proper} if it always
returns a hypothesis from $\H$. Throughout this paper we are concerned
with \emph{improper} learning, where the returned hypothesis can be any efficiently computed function $h$ from instances $\x$ to labels $y$.
Note that improper learning is just as useful as proper learning
for the purpose of deriving accurate predictors.


\subsection{A Warm-up Example}

To illustrate how more data can reduce runtime, consider the problem
of learning the class of $3$-term disjunctive normal form (DNF)
formulas in the realizable case. A $3$-DNF is a Boolean mapping, $h : \{0,1\}^d \to \{0,1\}$,
that can be written as $h(\x) = T_1(\x) \lor T_2(\x) \lor T_3(\x)$,
where for each $i$, $T_i(\x)$ is a conjunction of an arbitrary number
of literals, e.g. $T_i(\x) =x_1 \land \lnot x_3 \land x_5 \land \lnot
x_7$.

Since the number of $3$-DNF formulas is at most $3^{3d}$, it follows
that the information theoretic sample complexity is $O(d/\epsilon)$.
However, it was shown \cite{PittVa88,KearnsScSe94} that unless RP=NP,
the search problem of finding a $3$-DNF formula which is
(approximately) consistent with a given training set cannot be
performed in $\poly(d)$ time. On the other hand, we will show below
that if $m = \Theta(d^3/\epsilon)$ then $T_{\H,\epsilon}(m) = \poly(d/\epsilon)$.
Note that there is no contradiction between the last two sentences,
since the former establishes hardness of \emph{proper} learning while
the latter claims feasibility of improper learning.

To show the positive result, observe that each 3-DNF formula can be
rewritten as $\land_{u \in T_1,v \in T_2,w \in T_3} (u \lor v \lor w)$
for three sets of literals $T_1,T_2,T_3$.  Define $\psi : \{0,1\}^d
\to \{0,1\}^{2(2d)^3}$ such that for each triplet of literals $u,v,w$,
there are two indices in $\psi(\x)$, indicating if $u \lor v \lor w$
is true or false. Therefore, each 3-DNF can be represented as a single
conjunction over $\psi(\x)$. As a result, the class of $3$-DNFs over
$\x$ is a subset of the class of conjunctions over $\psi(\x)$. The
search problem of finding an ERM over the class of conjunctions is
polynomially solvable (it can be cast as a linear programming, or can
be solved using a simple greedy algorithm). However, the information
theoretic sample complexity of learning conjunctions over $2(2d)^3$
variables is $O(d^3/\epsilon)$.  We conclude that if $m =
\Theta(d^3/\epsilon)$ then $T_{\H,\epsilon}(m) = \poly(d/\epsilon)$.

It is important to emphasize that the analysis above is not
satisfactory for two reasons. First, we do not know if it is
not possible to improperly learn $3$-DNFs in polynomial time using $O(d/\epsilon)$
examples. All we know is that the ERM approach is not
efficient. Second, we do not know if the information theoretic sample
complexity of learning conjunctions over $\psi(\x)$ is
$\Omega(d^3/\epsilon)$. Maybe the specific structure of the range of
$\psi$ yields a lower sample complexity.

But, if we do believe that the above analysis indeed reflects reality,
we obtain two points on the curve $T_{\H,\epsilon}(m)$. Still, we do
not know how the rest of the curve looks like. This is illustrated
below.
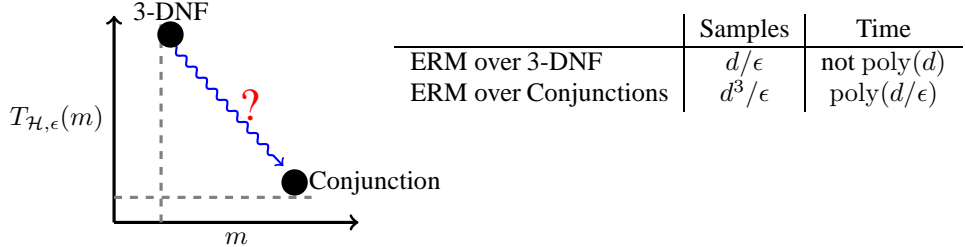
\begin{center}
\vskip -0.3cm
\begin{tikzpicture}[scale=2.5]

\draw [->,decorate,decoration={snake,amplitude=.4mm,segment
length=2mm,post length=1mm},blue,thick] (0.3,0.85)  -- node[right=1pt] {{\LARGE \color{red} ?}} (0.9,0.2);

\draw[->,very thick,black] (0,-0.1) --  node[left] {$T_{\H,\epsilon}(m)$} (0,1); \draw[->,very thick,black] (0,-0.1) -- node[below] {$m$}  (1.3,-0.1); \draw[dashed,very thick,gray] (0.25,-0.1) -- (0.25,1); \draw[dashed,very thick,gray] (0,100/3000) -- (1.05,100/3000);

\fill[color=black] (0.3,0.9) circle (0.07) node[above=2pt] {3-DNF}; \fill[color=black] (0.96000,0.11282) circle (0.07) node[right=2pt] {Conjunction};

\node at (3,0.75) {
\begin{tabular}{l|c|c}
& Samples & Time \\ \hline
ERM over 3-DNF & $d/\epsilon$ & not $\poly(d)$ \\ ERM over Conjunctions & $d^3/\epsilon$ & $\poly(d/\epsilon)$ \\ \end{tabular} };

\end{tikzpicture}

\end{center}

\section{Formal derivation of gaps}

In this section, we formally show a learning problem which exhibits an inverse dependence of the runtime on the number of examples. As discussed in the Subsection \ref{subsec:related}, it is distinguished from previous work in being applicable to the natural agnostic setting, where we do not assume that a perfect hypothesis exist. Since this assumption was crucial in all previous works, the construction we use is rather different.

To present the result, we will need the concept of a \emph{one-way permutation}. Intuitively, a one-way permutation over $\{0,1\}^n$ is a permutation which is computationally hard to invert. More formally, let $\U_n$ denote the uniform distribution over $\{0,1\}^n$, and let $\{0,1\}^{*}$ denote the set of all finite bit strings. Then we have the following definition:
\begin{definition}
A \emph{one-way permutation} $P:\{0,1\}^{*}\mapsto \{0,1\}^{*}$ is a function which for any $n$, maps $\{0,1\}^n$ to itself; there exists an algorithm for computing $P(\x)$, whose runtime is polynomial in the length of $\x$; and for any (possibly randomized) polynomial-time algorithm $A$ and any polynomial $p(n)$ over $n$, $\Pr_{\x\sim\U_n}(A(P(\x))=\x)< \frac{1}{p(n)}$ for sufficiently large $n$.
\end{definition}
It is widely conjectured that such one-way permutations exist. One concrete candidate is the RSA permutation function, which treats $\x\in \{0,1\}^n$ as a number in $\{0,\ldots,2^n-1\}$, and returns $P(\x)=\x^3~\text{mod}~N$, where $N$ is a product of two ``random'' primes of length $n$ such that $(p-1)(q-1)$ does not divide $3$. However, since the existence of such a one-way permutation would imply $P\neq NP$, there is no formal proof that such functions exist (see \cite{Goldreich01} for this and related results).

\begin{theorem}
There exists an agnostic binary classification learning problem over $\X=\{0,1\}^{2n}$ and $\Y=\{0,1\}$ with the following properties:
\begin{itemize}
    \item It is inefficiently learnable with sample size $m=O(1/\epsilon)$, and running time $O(2^n+m)$.
    \item Assuming one-way permutations exist, there exist no polynomial-time algorithm based on a sample of size $O(\log(n))$.
    \item It is efficiently learnable with a sample of size $m=O(n/\epsilon^2)$. Specifically, the training time is $O(m)$, resulting in an improper predictor whose runtime is $O(m^3)$.
 \end{itemize}
\end{theorem}

The theorem implies that in the reasonable regime where $1/\epsilon
\leq  \log(n)\leq n/\epsilon^2$, we really get an inverse dependence of the runtime on the training size. The theorem is illustrated below:
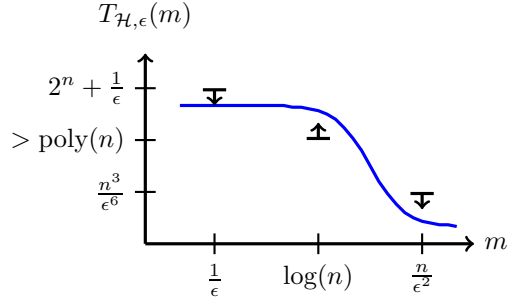
\begin{center}
\vskip -0.3cm
\begin{tikzpicture}[scale=2.3]

\draw[blue,very thick] (0.20,0.70) --(0.25,0.70) --(0.30,0.70) --(0.35,0.70) --(0.40,0.70) --(0.45,0.70) --(0.50,0.70) --(0.55,0.70) --(0.60,0.70) --(0.65,0.70) --(0.70,0.70) --(0.75,0.70) --(0.80,0.70) --(0.85,0.69) --(0.90,0.69) --(0.95,0.68) --(1.00,0.67) --(1.05,0.65) --(1.10,0.62) --(1.15,0.57) --(1.20,0.51) --(1.25,0.44) --(1.30,0.35) --(1.35,0.26) --(1.40,0.19) --(1.45,0.13) --(1.50,0.08) --(1.55,0.05) --(1.60,0.03) --(1.65,0.02) --(1.70,0.01) --(1.75,0.01) --(1.80,0.00) ;

\draw[|->,very thick,black] (0.4,0.8) -- (0.4,0.7);
\draw[|->,very thick,black] (1,0.5) -- (1,0.6);
\draw[|->,very thick,black] (1.6,0.2) -- (1.6,0.1);

\draw[->,very thick,black] (0,-0.1) --  (0,1) node[above=5pt]
{$T_{\H,\epsilon}(m)$};
\draw[-,thick,black] (0.06,0.8) -- (-0.06,0.8) node[left] {$2^n+\tfrac{1}{\epsilon}$};
\draw[-,thick,black] (0.06,0.5) -- (-0.06,0.5) node[left] {$>\poly(n)$};
\draw[-,thick,black] (0.06,0.2) -- (-0.06,0.2) node[left] {$\tfrac{n^3}{\epsilon^6}$};
\draw[->,very thick,black] (0,-0.1) --  (1.9,-0.1) node[right] {$m$};
\draw[-,thick,black] (1.6,-0.04) -- (1.6,-0.16) node[below] {$\tfrac{n}{\epsilon^2}$};
\draw[-,thick,black] (1,-0.04) -- (1,-0.16) node[below] {$\log(n)$};
\draw[-,thick,black] (0.4,-0.04) -- (0.4,-0.16)  node[below] {$\tfrac{1}{\epsilon}$};
\end{tikzpicture}
\end{center}
\vskip -0.4cm
To prove the theorem, we will define the following learning problem. Let $\X=\{0,1\}^{2n}$ and $\Y=\{0,1\}$. We will treat each $\x\in \X$ as a pair $(\r,\s)$, where $\r$ refers to the first $n$ bits in $\x$, and $\s$ to the last $n$ bits. Let $\inner{\r,\r'}=\sum_{i=1}^{n}r_i r'_i ~\text{mod}~2$ to denote inner product over the field $GF(2)$. Let $P$ be a one-way permutation. Then the example domain is the following subset of $\X\times \Y$:
\[
\Z = \{((\r,\s),b)~:~ \r,\s\in \{0,1\}^n, \inner{P^{-1}(\s),\r}=b\}.
\]
The loss function we use is simply the 0-1 loss, $\ell(h(\x),y)=\mathbf{1}_{h(\x)\neq y}$.

The hypothesis class $\H$ consists of randomized functions, parameterized by $\{0,1\}^n$, and defined as follows, where $\U_1$ is the uniform distribution on $\{0,1\}$:
\[
\H=\left\{h_{\x}(\r,\s)=\begin{cases} \inner{\x,\r} & \s=P(\x)\\ b\sim \U_1 & \text{otherwise}\end{cases} ~~:~~ \x\in \{0,1\}^n\right\},
\]

\subsubsection*{Learning with $O(\log(n))$ Samples is Hard}

We consider the following ``hard'' set of distributions $\{\D_{\x}\}$, parameterized by $\x\in \{0,1\}^n$: each $\D_{\x}$ is a uniform distribution over all $((\r,P(\x)),\inner{\x,\r})$. Note that there are exactly $2^n$ such examples, one for each choice of $\r\in \{0,1\}^n$. Also, note that for any such distribution $\D_{\x}$, $\inf_{h\in\H}\err(h)=0$, and this is achieved with the hypothesis $h_{\x}$.

First, we will prove that with a sample size $m=O(\log(n))$, any efficient learner fails on at least one of the distributions $\D_{\x}$. To see this, suppose on the contrary that we have an efficient distribution-free learner $A$, that works on all $\D_{\x}$, in the sense of seeing $m=O(\log(n))$ examples and then outputting some hypothesis $h$ such that $h((\r,P(\x)))=\inner{\x,\r}$ with even some non-trivial probability (e.g. at least $1/2+1/\poly(n)$). We will soon show how we can use such a learner $A$, such that in probability at least $1/\poly(n)$, we get an efficient algorithm $A'$, which given just $P(\x)$ and $r$, outputs $\inner{\x,\r}$ with probability at least $1/2+1/\poly(n)$. However, by the Goldreich-Levin Theorem (\cite{Goldreich01}, Theorem 2.5.2), such an algorithm can be used to efficiently invert $P$, violating the assumption that $P$ is a one-way permutation.

Thus, we just need to show how given $P(\x),\r$, we can efficiently compute $\inner{\x,\r}$ with probability at least $1/\poly(n)$. The procedure works as follows: we pick $m=O(\log(n))$ vectors $\r_1,\ldots,\r_m$ uniformly at random from $\{0,1\}^n$, and pick uniformly at random bits $b_1,\ldots,b_m$. We then apply our learning algorithm $A$ over the examples $\{((\r_i,P(\x)),b_i)\}_{i=1}^{m}$, getting us some predictor $h'$. We then attempt to predict $\inner{\x,\r}$ by computing $h'((\x',P(\x)))$.

To see why this procedure works, we note that with probability of $1/2^m = 1/\poly(n)$, we picked values for $b_1,\ldots,b_m$ such that $b_i=\inner{\x,\r_i}$ for all $i$. If this event happened, then the training set we get is distributed like $m$ i.i.d. examples from $\D_{\x}$. By our assumption on $A$, and the fact that $\inf_{h}\err(h)=0$, it follows that with probability at least $1/\poly(n)$, $A$ will return a hypothesis which predicts correctly with probability at least $1/2+1/\poly(n)$, as required.

\subsubsection*{Inefficient Distribution-Free Learning Possible with $O(1/\epsilon)$ Samples}

Ignoring computational constraints, we can use the following simple learning algorithm: given a training sample $\{(\r_i,\s_i),b_i\}_{i=1}^{m}$, find the most common value $\s'$ among $\s_1,\ldots,\s_m$, compute $\x'=P^{-1}(\s')$ (inefficiently, say by exhaustive search), and return the hypothesis $h_{\x'}$.

To see why this works, we will need the following lemma, which shows that if $h_{\x}$ has a low error rate, then $\s=P(\x)$ is likely to appear frequently in the examples (the proof appears in Appendix \ref{app:tech}).
\begin{lemma}\label{lem:probrelate}
For any distribution $\D$ over examples, and any fixed $\x\in \{0,1\}^n$, it holds that $Pr_{\s}(\s = P(\x)) = 1-2\err(h_{\x})$.
\end{lemma}

Suppose that $h_{\hat{\x}}$ is the hypothesis with a smallest generalization error in the hypothesis class. We now do a case analysis: if $\err(h_{\hat{\x}})> 1/2-\epsilon$, then the predictor $h_{\x'}$ returned by the algorithm is almost as good. This because the probability in the lemma statement cannot be negative, so for \emph{any} $\x'$ (and in particular the one used by the algorithm), we have $\err(h_{\x'})\leq 1/2$.

The other case we need to consider is that $\err(h_{\hat{\x}})\leq 1/2-\epsilon$. By the lemma, $\hat{\s}=P(\hat{\x})$
is the value of $\s$ most likely to occur in the sample (since $h_{\hat{\x}}$ is the one with smallest generalization error), and its probability of being picked is at least $1-2*(1/2-\epsilon) =\epsilon$. This means that after $O(1/\epsilon)$ examples, then with overwhelming probability, the $\s'$ we pick is such that $\Pr_{\s}(\s=\hat{\s})-\Pr_{\s}(\s=\s')\leq \epsilon/2$. But again by the lemma, this implies that $\err(h_{\x'})-\err(h_{\hat{\x}})$ is at most $\epsilon/4$. So $h_{\x'}$ that our algorithm returns is an $\epsilon/4$-optimal classifier as required.

\subsubsection*{Efficient Distribution-Free Learning Possible with $O(n/\epsilon^2)$ Samples}

We will need the following lemma, whose proof appears in Appendix \ref{app:tech}:
\begin{lemma}\label{lem:span}
Let $\D'$ be some distribution over $\{0,1\}^n$, and suppose we sample $m'$ vectors $\r_1,\ldots,\r_{m'}$ from that distribution. Then the probability that a freshly drawn vector $\r$ is not spanned by $\r_1,\ldots,\r_{m'}$ is at most $n/m'$.
\end{lemma}

We use a similar algorithm to the one discussed earlier for inefficient learning. However, instead of finding the most common $\s'$, computing $\x'=P^{-1}(\s')$ and returning $h_{\x'}$, which cannot be done efficiently, we build a predictor which is at most $\epsilon$ worse than $h_{\x'}$, and doesn't require us to find $\x'$ explicitly.

To do so, let $\{((\r_{i_j},\s_{i_j}),b_{i_j})\}_{j=1}^{m'}$ be the subset of examples for which $\s_{i_j}=\s'$. By definition of $\Z$, we know that for any such example, $\inner{\x',\r_{i_j}}=\inner{P^{-1}(\s'),\r_{i_j}}=b_{i_j}$. In other words, this gives us a set of values $\r_{i_1},\ldots,\r_{i_{m'}}$, for which we know $\inner{\x',\r_{i_1}},\ldots,\inner{\x',\r_{i_{m'}}}$. As a consequence, for any $\r$ in the linear subspace spanned by $\r_{i_1},\ldots,\r_{i_{m'}}$, we can efficiently compute $\inner{\x',\r}$. Let $B$ denote this subspace. Then our improper predictor works as follows, given some instance $(\r,\s)$:
\begin{itemize}
    \item If $\s=\s'$ and $\r\in B$, output $\inner{\x',\r}$ (note that this is the same output as $h_{\x'}$, by definition).
    \item If $\s\neq \s'$, output a random bit (note that this is the same output as $h_{\x'}$, by definition of $h_{\x'}$).
    \item If $\s=\s'$ and $\r\notin B$, output a bit uniformly at random.
\end{itemize}
Note that checking whether $\r\in B$ can always be done in at most $O(m'^3)\leq O(m^3)$ time, via Gaussian elimination.

Now, we claim that the probability of the third case happening is at most $\epsilon/2$. If this is indeed true, then our improper predictor is only $\epsilon/2$ worse (in terms of generalization error) from $h_{\x'}$, which based on the argument in the previous section, is already $\epsilon$-close to optimal.

So let us consider the possibility that $\s=\s'$ and $r\notin B$. If $\Pr_{\s}(\s=\s')\leq \epsilon$, we are done, so let us suppose that $\Pr_{\s}(\s=\s')> \epsilon$. This means that $m'$ is unlikely to be much smaller than $\epsilon m$. More precisely, by the multiplicative Chernoff bound, $\Pr(m'<\epsilon m/2) \leq \exp(-\epsilon m/8)$. Also, conditioned on some fixed $m'\geq \epsilon m/2$, \lemref{lem:span} assures us that $\Pr(\r\notin B|\s=\s')\leq n/m' \leq 2n/\epsilon m$. Overall, we get the following (the probabilities are over the draw of the training set and an additional example $((\r,\s),b)$):
\begin{align*}
&\Pr(\s=\s', \r\notin B) ~=~ \Pr(\s=\s',\r\notin B,m'<\epsilon m/2) + \Pr(\s=\s',\r\notin B, m'\geq \epsilon m/2)\\
&\leq \Pr(m'<\epsilon m/2)+\Pr(m'\geq \epsilon m/2, \r\notin B|\s=\s')\\
& \leq \exp(-\epsilon m/8)+\sum_{m'=\epsilon m/2}^{\infty}\Pr(m')\Pr(\r\notin B | \s=\s',m')~\leq~ \exp(-\epsilon m/8)+\frac{2n}{\epsilon m}.
\end{align*}
By taking $m=O(n/\epsilon^2)$ examples, we can ensure this to be at most order $\epsilon$.

\section{Gaps for natural learning problems}

In this section we collect examples of natural learning problems in
which we conjecture there is an inverse dependence of the training
time on the sample size. Some of these examples already appeared explicitly in previous literature, but most are new, unpublished, or did not appear in such an explicit form. We base our inverse dependence conjecture on the current best known upper bounds. Of course, an immediate open question is to show matching lower bounds. However, our main goal here is to demonstrate
general techniques of how to reduce the training runtime by requiring more
examples.



\subsection{Agnostically Learning Preferences}

Consider the set $[d] =\{1,\ldots,d\}$, and let $\X =
[d] \times [d]$ and $\Y = \{0,1\}$. That is, each
example is a pair $(i,j)$ and the label indicates whether $i$ is more
preferable to $j$.

Consider the hypothesis class of all permutations over $[d]$ which can
be written as $\H = \{ h_\w(i,j) = \indct{w_i > w_j} : \w \in
\reals^d\}$. The loss function is the 0-1 loss.  Note that each
hypothesis in $\H$ can be written as a Halfspace: $h_\w(i,j) =
\mathrm{sign}(\inner{\w,\e^i-\e^j})$.
Therefore, in the realizable case (namely, exists $h \in \H$ which
perfectly predicts the labels of all the examples in the training
set), solving the ERM problem can be performed in polynomial time.
However, in the agnostic case, finding a Halfspace that
minimizes the number of mistakes is in general NP hard. The sample
complexity of agnostically learning a $d$-dimensional Halfspace is
$\tilde{O}(d/\epsilon^2)$ and we therefore obtain that with a
non-efficient algorithm, it is possible to learn using
$\tilde{O}(d/\epsilon^2)$ examples.

On the other hand, in the following we show that with
$m=\Theta(d^2/\epsilon^2)$ it is possible to learn preferences in time
$O(m)$. The idea is to define the hypothesis class of all Boolean
functions over $\X$, namely, $H_1 = \{H(i,j) = M_{i,j} : M \in
\{0,1\}^{d^2} \}$. Clearly, $H \subset H_1$. In addition, $|H_1| =
2^{d^2}$ and therefore the sample complexity of learning $H_1$ using
the ERM rule is $O(d^2/\epsilon^2)$. Last, it is easy to verify that
solving the ERM problem can be easily done in time $O(m)$. So,
overall, we obtain the following:
\begin{center}
\vskip -0.3cm
%
%
%
%
%
%
\begin{tabular}{l|c|c}
& Samples & Time \\ \hline
ERM over $\H$ & $d/\epsilon^2$ & not $\poly(d)$ \\
ERM over $\H_1$ & $d^2/\epsilon^2$ & $d^2/\epsilon^2$ \\
\end{tabular}
\end{center}






\subsection{Agnostic Learning of Kernel-based Halfspaces}

We now consider the popular class of kernel-based linear
predictors. In kernel predictors, the instances $\x$ are mapped to a
high-dimensional feature space $\psi(\x)$, and a linear predictor is
learned in that space. Rather than working with $\psi(\x)$ explicitly,
one performs the learning implicitly using a kernel function
$k(\x,\x')$ which efficiently computes inner products
$\inner{\psi(\x),\psi(\x')}$ .

Since the dimensionality of the feature space may be high or even
infinite, the sample complexity of learning Halfspaces in the feature
space can be too large. One way to circumvent this problem is to
define a slightly different concept class by replacing the
non-continuous sign function with a Lipschitz continuous
function, $\phi : \reals \to [0,1]$, which is often called a transfer
function. For example, we can use a sigmoidal transfer function $
\phi_{\sig}(a) = 1/(1+ \exp(-4L\,a))$, which is a $L$-Lipschitz function. The resulting hypothesis class is $\H_\sig = \{\x \mapsto \phi_{\sig}(\inner{\w,\psi(\x)}) : \|\w\|_2 \le 1 \}$,
where we interpret the prediction $\phi_{\sig}(\inner{\w,\psi(\x)})
\in [0,1]$ as the probability to predict a positive label. The expected 0-1
loss then amounts to $\ell(\w,(\x,y)) = |y - \phi_{\sig}(\inner{\w,\psi(\x)})|$.

Using standard Rademacher complexity analysis (e.g. \cite{BartlettMe02}), it
is easy to see that the information theoretic sample complexity of
learning $\H$ is $O(L^2/\epsilon^2)$. However, from the computational
complexity point of view, the ERM problem amounts to solving a
non-convex optimization problem (with respect to $\w$). Adapting a
technique due to \cite{Ben-DavidSi00} it is possible to show that an
$\epsilon$-accurate solution to the ERM problem cam be calculated in
time $\exp\left(O\left(\tfrac{L^2}{\epsilon^2}
    \log(\tfrac{L}{\epsilon})\right)\right)$.  The idea is to observe
that the solution can be identified if someone reveals us a subset of
$(L/\epsilon)^2$ non-noisy examples. Therefore we can perform an
exhaustive search over all $(L/\epsilon)^2$ subsets of the $m$
examples in the training set and identify the best solution.

In \cite{ShalevShSr10}, a different algorithm has been proposed, that
learns the class $H_\sig$ using time and sample complexity
of at most $\exp\left(O\left(L\, \log(\tfrac{L}{\epsilon}
    )\right)\right)$. That is, the runtime of this algorithm is exponentially
smaller than the runtime required to solve the ERM problem using the technique
described in \cite{Ben-DavidSi00}, but the sample complexity is also
exponentially larger. The main idea of the algorithm given in
\cite{ShalevShSr10} is to define a new hypotheses class,
$\H_1 = \{ \x \mapsto \inner{\w,\hat{\psi}(\psi(\x))} : \|\w\|_2 \le B\}$,
where $B
= O((L/\epsilon)^L)$ and $\hat{\psi}$ is a mapping function for which
\[
\inner{\hat{\psi}(\psi(\x)),\hat{\psi}(\psi(\x'))} ~=~ \frac{2}{2-
  \inner{\psi(\x),\psi(\x')}} ~=~ \frac{2}{2- k(\x,\x')} ~.
\]
While it is not true that $\H \subset \H_1$, it is possible to show
that $\H_1$ ``almost'' contains $\H$ in the sense that for each $h \in
\H$ there exists $h_1 \in \H_1$ such that for all $\x$,
$|h(\x)-h_1(\x)| \le \epsilon$. The advantage of $\H_1$ over $\H$ is
that the functions in $\H_1$ are linear and hence the ERM problem with
respect to $\H_1$ boils down to a convex optimization problem and thus
can be solved in time $\poly(m)$, where $m$ is the size of the
training set. In summary, we obtain the following
\begin{center}
\begin{tabular}{l|c|c}
& Samples & Time \\ \hline
ERM over $\H$ & $L^2/\epsilon^2$ & $\poly\left(\exp\left(\tfrac{L^2}{\epsilon^2}
    \log(\tfrac{L}{\epsilon})\right)\right)$ \\
ERM over $\H_1$ & $\poly\left(\exp\left(L\, \log(\tfrac{L}{\epsilon}
    )\right)\right)$ & $\poly\left(\exp\left(L\, \log(\tfrac{L}{\epsilon}
    )\right)\right)$ \\
\end{tabular}
\end{center}

\subsection{Additional Examples}

In Appendix \ref{sec:additional} we list additional examples of
inverse dependence of runtime on sample size. These examples deal with
other learning settings like online learning and unsupervised
learning. These examples are interesting since they show other
techniques to obtain faster algorithms using a larger sample. For example, we demonstrate how to use \emph{exploration} for injecting structure into
the problem, which leads better runtime. The
price of the exploration is the need of a larger sample.  For the
unsupervised setting, we recall an existing example which shows
polynomial gap for learning the support of a certain sparse vector.

\section{Discussion}

In this paper, we formalized and discussed the phenomena of an inverse
dependence between the running time and the sample size. While this
phenomena has also been discussed in some earlier works, it was under a restrictive realizability assumption, that a perfect hypothesis exists, and the techniques mostly involved finding this hypothesis. In contrast, we frame our discussion in the more modern approach of agnostic and improper learning.

In the first half of our paper, we provided a novel construction which
shows such a tradeoff, based on a cryptographic assumption. While the
construction indeed has an inverse dependence phenomenon, it is not
based on a natural learning problem. In the second half of the paper,
we provided more natural learning problems, which seem to have this
phenomenon. Some of these problems were based on the intuition
described in the introduction, but some were based on other
techniques. However, the apparent inverse dependence in these problems is based on the assumption that the currently available upper bounds have matching lower bounds, which is not known to be true. Thus, we cannot formally prove that they indeed become computationally easier with the sample size.

Thus, a major open question is finding \emph{natural} learning problems, whose required running time has \emph{provable} inverse dependence with the sample size. We believe the examples we outlined hint at the existence of such problems, and provide clues as to the necessary techniques. Other problems are finding additional examples where this inverse dependence seems to hold, as well as finding additional techniques for making this inverse dependence happen. The ability to leverage large amounts of data to obtain more efficient algorithms would surely be a great asset to any machine learning application.

{\small
\bibliographystyle{plain}
\bibliography{mybib}
}

\newpage

\appendix

\section{Technical Results} \label{app:tech}

\subsection{Proof of \lemref{lem:probrelate}}

Using the definition of $\Z$ and $h_{\x}$, we have
\begin{align*}
&1-\err(h_{\x}) = \Pr_{((\r,\s),b)\sim \D}(b = h_{\x}(\r,\s)) \\
&= \Pr(\s=P(\x))~ \Pr(b = h_{\x}(\r,\s) |  \s=P(\x)) +
               \Pr(\s\neq P(\x)) ~ \Pr(b = h_{\x}(\r,\s) | \s\neq P(\x))\\
& = \Pr(\s=P(\x) ) * 1 + \Pr( \s\neq P(x) ) * \frac{1}{2} =
\frac{1}{2}( Pr( \s=P(\x) ) + 1 ).
\end{align*}
Rearranging, we get the result.

\subsection{Proof of \lemref{lem:span}}

Let $p_{k}$ denote the probability that after drawing $\r_1,\ldots,\r_{k}$, i.i.d., an independently drawn $\r_{k+1}$ is not spanned by $\r_1,\ldots,\r_{k}$. Also, let $B_{k}$ be a Bernoulli random variable with parameter $p_k$. Whenever $B_{k}=1$, the dimensionality of the subspace spanned by the vectors we drew so far increases by $1$. Since we are in an $n$-dimensional space, we must have $B_{1}+\ldots+B_{m'}\leq n$ with probability $1$. In particular, we have
\[
n \geq \E[B_{1}+\ldots+B_{m'}] = p_1+\ldots+p_{m'}.
\]
Also, for any $k\leq m'$, by the assumption that the vectors are drawn i.i.d., we have
\begin{align*}
p_m' &~=~ \Pr(r_{m'+1}\notin \text{span}(r_{1},\ldots,r_{m'}))~\leq~ \Pr(r_{m'+1} \notin\text{span}(r_{1},\ldots,r_{k})) \\&~=~ \Pr(r_{k+1}\notin \text{span} (r_{1},\ldots,r_{k})) ~=~ p_k.
\end{align*}
Combining the two inequalities, it follows that $m'p_{m'}\leq n$, so $p_{m'}\leq n/m'$ as required.

\section{Additional Examples} \label{sec:additional}

\subsection{Online Multiclass Categorization with Bandit Feedback}

This example is based on \cite{KakadeShTe08a}. It deals with another
variant of the multi-armed bandit problem. It shows how to use
\emph{exploration} for injecting structure into the problem, which
leads to a decrease in the required runtime. The price of the
exploration is a larger regret, which corresponds to the need of a
larger number of online rounds for achieving the same target error.

The setting is as follows. At each online round, the
learner first receives a vector $\x_t \in \reals^d$ and need to
predict one of $k$ labels (corresponding to arms). Then, the
environment picks the correct label $y_t$, without revealing it to the
learner, and only tells the learner the binary feedback of if his
prediction was correct or not.

We analyze the number of mistakes the learner will perform in $T$
rounds, where we assume that there exists some matrix $W^\star \in \reals^{k,d}$ such
that at each round the correct label is $y_t = \arg\max_{y \in [k]}
(W^\star \x)_y$. We further assume that the score of the correct label
is higher than the runner-up by at least $\gamma$ and that the
Frobenius norm of $W^\star$ is at most $1$. We also assume that $d$ is
order of $1/\gamma^2$ (this is not restricting due to the possibility
of performing random projections).

We now consider two algorithms. The first uses a multiclass version of
the Halving algorithm (see \cite{KakadeShTe08a}) which can be
implemented with the bandit feedback and has a regret bound of
$\tilde{O}(k^2 d)$. However, the runtime of this algorithm is  $2^{kd}$.

The second algorithm is the Banditron of \cite{KakadeShTe08a}. The
Banditron uses exploration for reducing the learning problem into the
problem of learning multiclass classifier in the full information
case, which can be performed efficiently using the Perceptron
algorithm. In particular, in some of the rounds the Banditron guesses
a random label, attempting to ``fish'' the relevant information. This
exploration yields a higher regret bound of $O(\sqrt{k d T})$.

We can therefore draw the following table, which shows a tradeoff
between running time and number of rounds required to obtain regret
$\leq \epsilon$. Note that we will usually want $\epsilon$ to be much
smaller than $1/k$.

\begin{center}
\begin{tabular}{l|c|c}
& Rounds & Time \\ \hline
Inefficient alg. & $k^2d/\epsilon$ & $T 2^{kd}$ \\
Efficient alg. & $ kd/\epsilon^2$ & $Tkd$\\
\end{tabular}
\end{center}

\subsection{Sparse Principal Component Recovery}

This example is taken from \cite{Wainwright09}. This time, it is in the context of \emph{unsupervised} statistical learning.

The problem is as follows: we have an i.i.d. sample of vectors drawn from  $\reals^d$. The distribution is assumed to be Gaussian $\mathcal{N}(\mathbf{0},\Sigma)$, with a ``spiked'' covariance structure. specifically, the covariance matrix $\Sigma$ is assumed to be of the form $I_{d}+\z \z^{\top}$, where $\z$ is an unknown \emph{sparse} vector, with only $k$ non-zero elements of the form $\pm 1/\sqrt{k}$. Our goal in this setting is to detect the support of $\z$.

\cite{Wainwright09} provide two algorithms to deal with this problem. The first method is a simple diagonal thresholding scheme, which takes the empirical covariance matrix $\hat{\Sigma}$, and returns the $k$ indices for which the diagonal entries of $\hat{\Sigma}$ are largest. It is proven that if $m\geq c k^2\log(d-k)$ (for some constant $c$), then the probability of not perfectly identifying the support of $\z$ is at most $\exp(-O(k^2\log(d-k)))$, which goes to $0$ with $k$ and $d$. Thus, we can view the sample complexity of this algorithm as $O(k^2\log(d-k))$. In terms of running time, given a sample of size $m=O(k^2\log(d-k))$, the method requires computing the diagonal of $\hat{\Sigma}$ and sorting it, for a total runtime of $O(k^2d\log(d-k)+d\log(d)) = O(k^2 d\log(d))$.

The second algorithm is a more sophisticated semidefinite programming (SDP) scheme, which can be solved exactly in time $O(d^4\log(d))$. Moreover, the sample complexity for perfect recovery is shown to be asymptotically $O(k\log(d-k))$. Summarizing, we have the following clear sample-time complexity tradeoff. Note that here, the gaps are only polynomial.

\begin{center}
\begin{tabular}{l|c|c}
& Samples & Time \\ \hline
SDP & $k\log(d-k)$ & $d^4\log(d)$ \\
Thresholding & $k^2\log(d-k)$ & $k^2d\log(d)$\\
\end{tabular}
\end{center}

\end{document}